\documentclass[acmtog, authorversion, nonacm]{acmart} %

\usepackage{booktabs} %
\usepackage{cleveref}
\usepackage{graphicx}
\usepackage{booktabs}
\usepackage{wrapfig,lipsum,booktabs}
\usepackage{multirow}
\usepackage{enumitem} %

\usepackage[ruled]{algorithm2e} %

\SetAlFnt{\small}
\SetAlCapFnt{\small}
\SetAlCapNameFnt{\small}
\SetAlCapHSkip{0pt}

\acmJournal{TOG}

\newif\ifdraft
\drafttrue

\ifdraft
\newcommand{\sagie}[1]{{\color{red}[\textbf{Sagie:} #1]}}
\newcommand{\yoel}[1]{{\color{orange}[\textbf{Yoel:} #1]}}
\newcommand{\david}[1]{{\color{green}[\textbf{David:} #1]}}
\newcommand{\itai}[1]{{\color{cyan}[\textbf{Itai:} #1]}}

\else
\newcommand{\sagie}[1]{}
\newcommand{\yoel}[1]{}
\newcommand{\david}[1]{}
\newcommand{\itai}[1]{}

\fi

\usepackage{amsthm}
\usepackage{amsmath,amsfonts,bm}
\makeatletter
\newtheorem*{rep@theorem}{\rep@title}
\newcommand{\newreptheorem}[2]{%
\newenvironment{rep#1}[1]{%
 \def\rep@title{#2 \ref{##1}}%
 \begin{rep@theorem}}%
 {\end{rep@theorem}}}
\makeatother

\newreptheorem{theorem}{Theorem}

\definecolor{myred}{RGB}{215,48,39}
\definecolor{mygreen}{RGB}{26,152,80}

\newcommand{\halfmark}{\textcolor{gray}{\checkmark\kern-1.1ex\raisebox{.7ex}{\rotatebox[origin=c]{125}{--}}}}
\usepackage[framemethod=TikZ]{mdframed}
\mdfdefinestyle{MyFrame}{%
    linecolor=black,
    outerlinewidth=.3pt,
    roundcorner=5pt,
    innertopmargin=1pt, %
    innerbottommargin=1pt, %
    innerrightmargin=1pt,
    innerleftmargin=1pt,
    backgroundcolor=black!0!white}

\mdfdefinestyle{MyFrame2}{%
    linecolor=white,
    outerlinewidth=1pt,
    roundcorner=1pt,
    innertopmargin=0,
    innerbottommargin=0,
    innerrightmargin=7pt,
    innerleftmargin=7pt,
    backgroundcolor=black!3!white}

\mdfdefinestyle{MyFrameEq}{%
    linecolor=white,
    outerlinewidth=0pt,
    roundcorner=0pt,
    innertopmargin=0pt,
    innerbottommargin=0pt,
    innerrightmargin=7pt,
    innerleftmargin=7pt,
    backgroundcolor=black!3!white}

\newcommand{\RNum}[1]{\uppercase\expandafter{\romannumeral #1\relax}}

\newcommand{\vertiii}[1]{{\left\vert\kern-0.25ex\left\vert\kern-0.25ex\left\vert #1 
    \right\vert\kern-0.25ex\right\vert\kern-0.25ex\right\vert}}
\newcommand{\vertiiii}[1]{{\vert\kern-0.25ex\vert\kern-0.25ex\vert #1 
    \vert\kern-0.25ex\vert\kern-0.25ex\vert}}

\usepackage{mathtools}

\newcommand{\cut}[1]{}

\newcommand{\removelatexerror}{\let\@latex@error\@gobble}

\def\eqref#1{Eq.~\ref{#1}}

\def\1{\bm{1}}

\DeclareMathAlphabet{\mathsfit}{\encodingdefault}{\sfdefault}{m}{sl}
\SetMathAlphabet{\mathsfit}{bold}{\encodingdefault}{\sfdefault}{bx}{n}

\newcommand{\sethree}{\mathrm{SE(3)}}

\newcommand{\cfm}[1]{$\sethree$-CFM\xspace}
\newcommand{\otcfm}[1]{$\sethree$-OT-CFM\xspace}
\newcommand{\sfm}[1]{$\sethree$-SFM\xspace}
\newcommand{\foldflow}[1]{\textsc{FoldFlow}\xspace}
\newcommand{\foldflowbase}[1]{\textsc{FoldFlow-Base}\xspace}
\newcommand{\foldflowot}[1]{\textsc{FoldFlow-OT}\xspace}
\newcommand{\foldflowsfm}[1]{\textsc{FoldFlow-SFM}\xspace}

\begin{document}

\title{Structurally Disentangled Feature Fields Distillation for 3D Understanding and Editing}

\author{Yoel Levy}
\affiliation{%
 \institution{The Hebrew University of Jerusalem}
 \country{Israel}
 }
 \author{David Shavin}
\affiliation{%
 \institution{The Hebrew University of Jerusalem}
 \country{Israel}
 }
  \author{Itai Lang}
\affiliation{%
 \institution{University of Chicago}
 \country{United States of America}
 }
  \author{Sagie Benaim}
\affiliation{%
 \institution{The Hebrew University of Jerusalem}
 \country{Israel}
 }

\begin{abstract}

Recent work has demonstrated the ability to leverage or distill pre-trained 2D features obtained using large pre-trained 2D models into 3D features, enabling impressive 3D editing and understanding capabilities using only 2D supervision. 
Although impressive, the input 2D features may be 3D and physically inconsistent. Current models average such inconsistencies, resulting in inferior distilled 3D features. 
In this work, we propose instead to capture 3D features using multiple disentangled feature fields that capture different structural components of 3D features involving view-dependent and view-independent components. 
Subsequently, each element can be controlled in isolation, enabling semantic and structural understanding and editing capabilities. For instance, 3D segmentation can be achieved by considering only view-indepenent features, and discarding the view-dependent ones, resulting in significant performance compared to baselines, which average view-dependent features. Using a user click, one can segment 3D features corresponding to a given object and then segment, edit, or remove their view-dependent (reflective) properties, enabling 
a set of novel understanding and editing tasks.

\end{abstract}



\begin{teaserfigure}
\centering
\begin{tabular}{@{}c@{}c@{}c@{}c@{}c@{}c@{}}
Novel View & Sphere Segmentation & Refl. Segmentation & Color Change & Rougher & W/o Refl. \\ 
\includegraphics[width=0.165\linewidth]{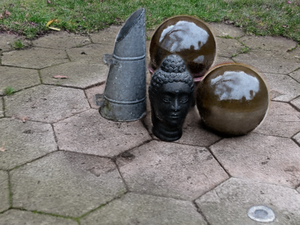} &
\includegraphics[width=0.165\linewidth]{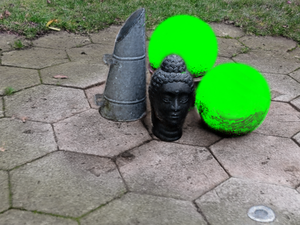} &
\includegraphics[width=0.165\linewidth]{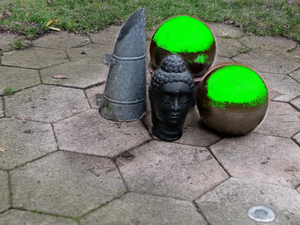} &
\includegraphics[width=0.165\linewidth]{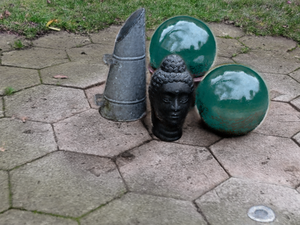} &
\includegraphics[width=0.165\linewidth]{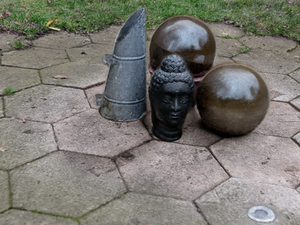} &
\includegraphics[width=0.165\linewidth]{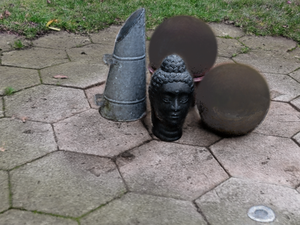} \\
\includegraphics[width=0.165\linewidth]{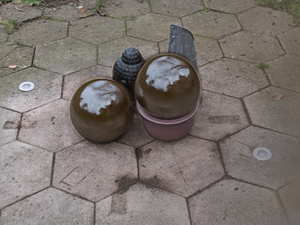} &
\includegraphics[width=0.165\linewidth]{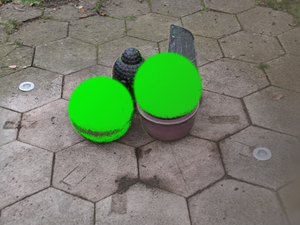} &
\includegraphics[width=0.165\linewidth]{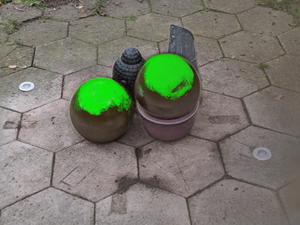} &
\includegraphics[width=0.165\linewidth]{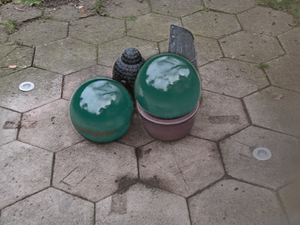} &
\includegraphics[width=0.165\linewidth]{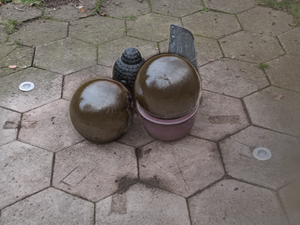} &
\includegraphics[width=0.165\linewidth]{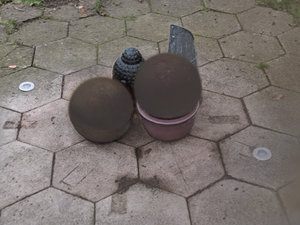} \\
\end{tabular}
\vspace{-0.3cm}
\caption{An illustration of our method for the Garden-spheres scene from the real-world RefNeRF  dataset~\cite{verbin2022refnerf}. By distilling 2D DINOv2~\cite{oquab2023dinov2} features into our disentangled feature field representation, our method enables: (i). Generation of novel views, (ii). 3D segmentation of the spheres, (iii). 3D segmentation of the reflective region of the spheres, (iv). Color editing while adhering to reflections, (v). Changing the roughness of the spheres, (vi). Removing the reflection from the spheres (using both the reflection and camera-based components). Refl. = Reflection.}
\label{fig:teaser}
\end{teaserfigure}

\maketitle

\section{Introduction}
\label{sec:intro}

A recent paradigm in computer graphics and computer vision is 2D feature distillation, whereby 2D image features obtained using large-scale self-supervised methods are ``distilled" to the parameters of an underlying 3D model. Doing so enables 3D semantic understanding and editing, given 2D RGB and feature supervision only. Several works~\cite{kobayashi2022decomposing, kerr2023lerf, ye2023featurenerf} considered the setting of novel view synthesis with an underlying NeRF~\cite{mildenhall2021nerf} or a Gaussian Splatting~\cite{kerbl20233dgd} model, achieving impressive 3D understanding and editing capabilities. Each 3D point (in the former) or Gaussian (in the latter) stores a feature value, which can be rendered to different views using volumetric rendering. The rendered views correspond to 2D feature maps obtained using self-supervised 2D methods such as DINOv2~\cite{oquab2023dinov2}. This paradigm was later extended to text~\cite{kerr2023lerf, qin2023langsplat}, enabling impressive 3D text-based understanding and editing capabilities.
Although impressive, current models assume that 3D features are captured using a single view-independent feature field (or a single 3D feature value per point or Gaussian). However, as demonstrated in Sec.~\ref{sec:feature_analysis}, input 2D features from SSL models contain 3D inconsistencies and are inherently view dependent (i.e., contain significant view-dependent variations). Current models average such view-dependent variations, resulting in inferior features for downstream performance on tasks such as 3D segmentation. 

In this work, we propose instead to capture 3D features using \textit{multiple disentangled feature fields} that capture different structural components of 3D features. While these structural components could involve a variety of physical properties such as lighting and deformations, we focus on the disentanglement of view-dependent and view-independent features. 
Our disentangled feature fields can be learned using 2D supervision only, in an unsupervised manner, thus enabling the disentanglement of 2D (and 3D) features into components that are view-independent and view-dependent (reflections and camera-dependent features). Subsequently, each component can be controlled separately, enabling semantic and structural understanding and editing capabilities. 
For instance, 3D segmentation can be achieved by considering only view-indepenent features, and discarding the view-dependent ones, resulting in significant performance compared to baselines, which average view-dependent features. 
Using a user click, one can segment an entire object in 3D or only its reflective component, edit view-independent object properties, such as its color, while adhering to reflections, or remove the object's reflections. An illustration is provided in Fig.~\ref{fig:teaser}.

To achieve our desired disentanglement, we propose computing the feature value of a 3D point along a viewing direction as the combination of the outputs of two disentangled feature fields: (i). A reflected view feature field capturing view-dependent features that arise from specular object reflections, and (ii). A view-independent feature field capturing diffuse features of the object which depend only on the location of a 3D point and not the viewing direction.

We evaluate our approach on a variety of objects from a diverse set of scenes from the Shiny Blender dataset \cite{verbin2022refnerf} as well as real-world scenes from the RefNeRF real-world dataset \cite{verbin2022refnerf} and Mip-NeRF-360 dataset \cite{barron2022mipnerf360}.
In terms of 3D understanding, we consider the tasks of: (i). 3D segmentation, for which our structurally disentangled representation achieves superior results compared to a single holistic feature field \cite{kobayashi2022decomposing},
and (ii). Segmentation of view-dependent components in a scene. For example, one can segment only the reflective component of an object selected using a user click.  In terms of 3D editing, we demonstrate the applicability of our approach on (i). The ability to remove the reflective component of an object, and lastly (ii). The ability to edit individual 3D component. For example, changing the object's color while correctly adhering to reflections or manipulating its roughness.

In summary, we offer the following contributions:
\begin{enumerate}[nolistsep]
\item To our knowledge, we are the first to consider a physically-inspired decomposition of a \textit{semantic} feature field. This is in contrast to previous works, such as RefNeRF \cite{verbin2022refnerf} and UniSDF \cite{wang2023unisdf}, which only consider appearance, and to DFF \cite{kobayashi2022decomposing}, which uses a single view-independent feature field. Our decomposition reveals a new scientific insight: the view-dependent features of a 2D foundation model \textit{can} be decomposed into a 3D feature field of view-independent and view-dependent components.
\item By doing so, we can: (i) Introduce novel applications, including view-dependent and view-independent segmentation and editing capabilities. (ii) Improve 3D segmentation by only using the view-independent component and discarding the view-dependent component, demonstrating state-of-the-art performance.
\item We improve the underlying reference features of the foundation model by \textit{decomposing} the view-dependent component. In contrast, baseline models such as DFF \textit{averages} all the view-dependent feature components by modeling only a single view-independent feature field.
\end{enumerate}

\section{Related Work} \label{sec:related}

\subsection{Structured Novel View Synthesis}

Neural Radiance Field (NeRF)~\cite{mildenhall2021nerf} reconstructs a 3D scene from 2D images by mapping 3D spatial locations and viewing directions to their corresponding color and density values. %
Different works introduced physical modeling within a NeRF-like formulation to allow for the modeling of geometry, lighting, materials, reflections, and more. 
For instance, to model geometry, many works integrate a signed distance function (SDF) formulation~\cite{li2023neuralangelo, long2022sparseneus, oechsle2021unisurf, rosu2023permutosdf, wang2021neus, yariv2021volume, yu2022monosdf}. 
Other works extended NeRF to enable relighting by disentangling appearance into scene lighting and materials~\cite{bi2020neural, boss2021nerd, boss2021neural, srinivasan2021nerv, zhang2021physg, zhang2021nerfactor}.

In the context of reflections modeling,  approaches such as \cite{ramamoorthi2009precomputation} represented appearance using disentangled view-dependent specular and reflective appearance. 
Ref-NeRF~\cite{verbin2022refnerf} modeled appearance through two disentangled radiance fields within an implicit radiance field formulation, one modeling view-independent diffuse properties and another modeling reflective properties.
BakedSDF~\cite{yariv2023bakedsdf} and  ENVIDR~\cite{liang2023envidr} extended this representation for the reconstruction of glossy surfaces and with material decomposition. UniSDF~\cite{wang2023unisdf} improved upon these parameterizations by including two separate radiance fields for modeling reflections, one for the reflected view and another for camera view based radiance. 
Unlike the above, our focus is on structural disentanglement for feature fields, modeling 3D features together with appearance. As far as we can ascertain, our work is the first to enable such disentanglement.

\subsection{2D Feature Distillation}

Due to the scarcity of 3D data, recent work attempted to distill or lift 2D features trained using pre-trained self-supervised 2D models (such as DINO~\cite{caron2021emerging}, DINOv2~\cite{oquab2023dinov2} or CLIP-LSeg~\cite{li2022languagedriven}). In particular, \cite{kobayashi2022decomposing, tschernezki2022neural} extended NeRF to model not only appearance, but also semantics, through the use of an additional view-independent feature field, that maps a 3D point into a 3D feature. 
Other works embedded semantic information into NeRFs~\cite{siddiqui2023panoptic, yao2024matte}. Further works focused on integrating open-vocabulary text-based features, by embedding CLIP features into NeRF~\cite{kerr2023lerf} or Gaussian splatting~\cite{qin2023langsplat}. 
Although impressive, models assume that 3D features are captured using a single view-independent feature field (or a single 3D feature per point or Gaussian)
Our work instead focuses on the 3D model, capturing 3D features using multiple structurally disentangled feature fields, 
thus enabling multiple novel understanding and editing capabilities.

\section{Method} \label{sec:method}

\subsection{Preliminaries}

NeRF~\cite{mildenhall2021nerf} represents a 3D scene using a multi-layer perceptron network (MLP) that parameterizes a continuous 5D radiance field. This field $f:(\mathbf{x}, \mathbf{d}) \rightarrow (\mathbf{c}, \sigma)$, maps a 3D coordinate $\mathbf{x} \in \mathbb{R}^3$ and a viewing direction $\mathbf{d} \in \mathbb{R}^2$ to an emitted color $\mathbf{c} \in \mathbb{R}^3$ and volume density $\sigma \in \mathbb{R}$. 
To render 2D views, NeRF employs volumetric rendering. In particular, rays are cast through the scene, with $f(\mathbf{x}, \mathbf{d})$ queried along each ray. The volume rendering equation is then used to composite color and opacity:
\begin{align}
C(r) = \int T(t) \mathbf{c} (r(t)) \sigma(r(t)) dt,
\end{align}

\noindent where $r(t)$ parameterizes the ray, $T(t)$ is the accumulated transmittance, and $\mathbf{c}(r(t))$ and $\sigma(r(t))$ are the color and density at point t.
NeRF is trained by minimizing a reconstruction loss between images rendered from the field and a set of ground-truth images with known camera poses. This differentiable process allows NeRF to implicitly learn scene geometry and appearance for photorealistic rendering from unseen viewpoints.

When considering feature distillation~\cite{kobayashi2022decomposing, kerr2023lerf, ye2023featurenerf}, one additionally learns an MLP that parameterizes a continuous 3D feature field $f_{feat}$. Unlike $f$, $f_{feat} : \mathbf{x} \rightarrow \mathbf{f}$ maps a 3D coordinate $\mathbf{x} \in \mathbb{R}^3$ to a feature value $\mathbf{f} \in \mathbb{R}^n$. 2D feature maps can then be rendered similarly to color values, using $\sigma$ predicted by $f$. That is:
\begin{align}
F(r) = \int T(t) \mathbf{f}(r(t)) \sigma(r(t)) dt,
\end{align}

\noindent where $r(t)$, $T(t)$ and $\sigma(r(t))$ are as above and $\mathbf{f}(r(t))$ represents the predicted feature value at point $t$ (identical for every view direction). In addition to image-based reconstruction loss, one also minimizes a reconstruction loss between rendered features and ground-truth features, obtained by applying a pre-trained 2D feature extractor (such as DINOv2~\cite{oquab2023dinov2}) to ground-truth RGB views.

\subsection{Structurally Disentangled Feature Fields}
\begin{figure}[t!]
\centering
\includegraphics[width=\linewidth]{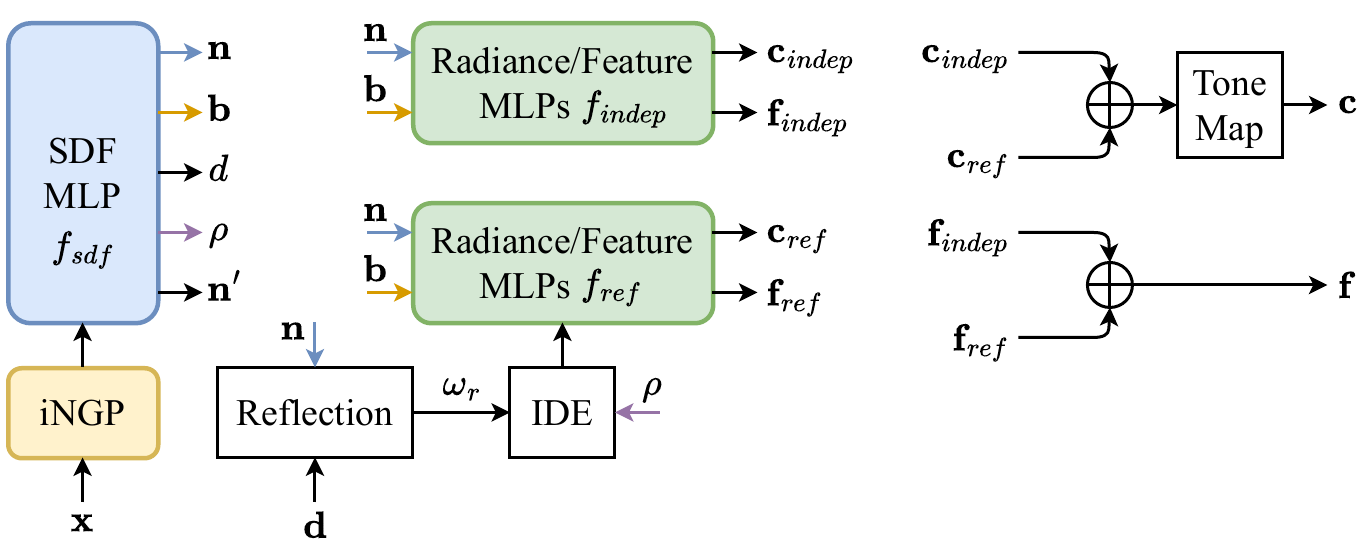}
\vspace{-0.7cm}
\caption{\textbf{The method's pipeline.} We decompose the appearance color of the scene $\mathbf{c}$ into physical components $\mathbf{c}_{indep}$ and $\mathbf{c}_{ref}$ and sum them to compute the color of the scene at location $\mathbf{x}$ and viewing direction $\mathbf{d}$. We also learn a decomposed feature field for the scene, $\mathbf{f}_{indep}$ and $\mathbf{f}_{ref}$, which enables physically-oriented semantic understanding and editing applications. Please see \cref{sec:method} for more details.
} 
\label{fig:pipeline}
\vspace{-0.3cm}
\end{figure}

Fig.~\ref{fig:pipeline} illustrates our method, which we elaborate on below. 

\smallskip
\noindent \textbf{Signed distance representation.} Similar to previous work~\cite{wang2023unisdf}, our model's backbone is an MLP computing the signed distance function $d(\mathbf{x})$ (SDF) at each location $\mathbf{x}$. The scene's surface is defined at the zero level set of the function. 
$\mathbf{x}$ undergoes a contraction transformation~\cite{barron2022mipnerf360, wang2023unisdf}, limiting its values to $[0, 2)$.

The density $\sigma(\mathbf{x})$ for volume rendering is computed by $\sigma(\mathbf{x}) = \alpha \Psi_\beta(d(\mathbf{x}))$, where $\Psi_\beta$ is the cumulative distribution of a Laplace distribution with zero mean and a scale parameter $\beta$ that is learned during optimization. The benefit of the SDF representation is that the normal of the scene surface can be easily derived as the gradient of the distance function:
\begin{align} \label{eq:normal}
\mathbf{n}(\mathbf{x}) = \nabla d(\mathbf{x}) / || \nabla d(\mathbf{x}) ||_2.
\end{align}

\noindent The normal is used as input to the components in our model for computing the disentangled feature representation. 

To better reconstruct high-frequency details and to speed up training, we use the iNGP hash encoding~\cite{muller2022instant}. Each location $\mathbf{x}$ is mapped to a high-dimensional feature vector, which is the concatenation of the iNGP's pyramid-level features. This feature vector is the input to our SDF MLP. To ease notations, we omit in this section the notation of $\mathbf{x}$ for location-dependent quantities.

\smallskip
\noindent \textbf{Radiance and features components.} We decompose the appearance of the scene into two elements: a view-independent color $\mathbf{c}_{indep}$ and a view-dependent reflected color $\mathbf{c}_{ref}$. The view-independent color is calculated as:
\begin{align} \label{eq:c_indep}
\mathbf{c}_{indep} = f_{indep}(\mathbf{n}, \mathbf{b}),
\end{align}

\noindent where $f_{indep}$ is a learned MLP, $\mathbf{n}$ is the normal defined in Eq.~\ref{eq:normal}, and $\mathbf{b}$ is a bottleneck vector output from the SDF MLP. This vector enables additional degrees of freedom to accommodate second-order effects, such as varying illumination over the scene~\cite{verbin2022refnerf, wang2023unisdf}. $\mathbf{c}_{indep}$ represents the view-independent color and accordingly is independent of the viewing direction $\mathbf{d}$.

The other color component, $\mathbf{c}_{ref}$, is view dependent. It is responsible for reflectivity scene regions and models the reflected radiance in the scene. Following previous work~\cite{verbin2022refnerf}, we calculate the reflection for the viewing direction about the normal:
\begin{align} \label{eq:omega_r}
\mathbf{\omega}_r = 2(\mathbf{\omega}_o \cdot \mathbf{n}) \mathbf{n} - \mathbf{\omega}_o,
\end{align}

\noindent where $\mathbf{\omega}_o = -\mathbf{\hat{d}}$, is a unit vector pointing to the camera from a point in
space, and $\mathbf{\hat{d}}$ is a viewing direction. Then, we use Integrated Directional Encoding (IDE) \cite{verbin2022refnerf} in the computation of $\mathbf{c}_{ref}$:
\begin{align} \label{eq:reflectence_color}
\mathbf{c}_{ref} = f_{ref}(\mathbf{n}, \mathbf{b}, \text{IDE}(\mathbf{\omega}_r, \kappa)),
\end{align}
\noindent where $\kappa = 1/\rho$, and $\rho$ is the surface roughness predicted by the SDF MLP. 
Finally, the rendered color is given by:
\begin{align}
\mathbf{c} = \text{tonemap}(\mathbf{c}_{indep} + \mathbf{c}_{ref}),
\label{eq:learnable_weights_1}
\end{align}

\noindent where $\text{tonemap}(\cdot)$ is a tone mapping function converting linear color to the sRGB format and clipping the output to the range $[0, 1]$. Similarly, we decompose the feature field of the scene $\mathbf{f}$ as follows:
\begin{align}
\mathbf{f} = \mathbf{f}_{indep} + \mathbf{f}_{ref},
\label{eq:learnable_weights_2}
\end{align}

\noindent where $\mathbf{f}_{indep}$ and $\mathbf{f}_{ref}$ are computed in the same manner as $\mathbf{c}_{indep}$, Eq.~\ref{eq:c_indep} and $\mathbf{c}_{ref}$, Eq.~\cref{eq:omega_r,eq:reflectence_color}. The feature components are used as a \textit{semantic} representation for their color counterparts. We note that the semantic feature from the independent feature field only, $\mathbf{f}_{indep}$ can be used for understanding tasks such as 3D segmentation, as 3D segmentation is inherently view-independent. This is an example where a physical disentanglement (i.e., to view-dependent and independent components) can result in ``improving" or ``cleaning" undesirable feature components required for 3D segmentation. By jointly modeling features and colors, one can also enable a diversity of editing applications as illustrated in Sec.~\ref{sec:results}.

\subsection{Training Objective}
\label{sec:training_objective}

The training of our structural decomposition of the scene is supervised by posed images and their corresponding DINOv2~\cite{oquab2023dinov2} feature maps. Given a set of posed images $\{C^{gt}_1, \dots, C^{gt}_m\}$, where $C^{gt}_i \in \mathbb{R}^{H \times W \times 3}$, we obtain associated 2D feature maps $\{F^{gt}_1, \dots, F^{gt}_m\}$ using the pretrained feature extractor DINOv2, where $F^{gt}_i \in \mathbb{R}^{H \times W \times n}$, and $n$ is the feature dimension.
The rendered color images are compared to the given images with an $l_2$ loss:
\begin{align}
\mathcal{L}_c = \frac{1}{m} \sum_i { || C_i - C^{gt}_i ||_2^2 },
\end{align}

\noindent where $C_i$ is the scene appearance rendered from our model for the view direction of image $C^{gt}_i$. Additionally, we regularize the SDF MLP learning with the eikonal loss~\cite{icml2020_2086, wang2023unisdf} to promote the approximation of a valid SDF:
\begin{align}
\mathcal{L}_{e} = \frac{1}{| \mathbf{x} |} \sum_{\mathbf{x}} {(|| \nabla \mathbf{d}(\mathbf{x}) ||_2 - 1)^2}.
\end{align}

We also encourage normal smoothness by comparing the computed normal $\mathbf{n}(\mathbf{x})$ from the SDF to the normal $\mathbf{n'}(\mathbf{x})$ predicted by the SDF MLP:
\begin{align}
\mathcal{L}_{p} = \sum_{\mathbf{x}} w_\mathbf{x} \: {|| \mathbf{n}(\mathbf{x}) - \mathbf{n'}(\mathbf{x})||^2}.
\end{align}

Further, we penalize back-facing normals using the orientation loss \cite{verbin2022refnerf}:
\begin{align}
\mathcal{L}_{o} = \sum_{\mathbf{x}} w_\mathbf{x} \: {\max(0, \mathbf{n}(\mathbf{x}) \cdot \mathbf{d}(\mathbf{x}))^2}.
\end{align}

For feature learning, we have found that optimizing the radiance and feature fields concurrently compromises the learning process of both. Thus, we first learn the decomposed appearance of the scene using the total loss:
\begin{align}
\mathcal{L} = \mathcal{L}_c + \lambda_e \mathcal{L}_e + \lambda_p \mathcal{L}_p + \lambda_o \mathcal{L}_o.
\end{align}

\noindent Then, we freeze the appearance model and train MLPs for optimizing our feature field decomposition using: 
\begin{align}
\mathcal{L}_f = \frac{1}{m} \sum_i { || F_i - F^{gt}_i ||_2^2 }.
\label{eq:full_objective}
\end{align}

\subsection{Segmentation and Editing}
\label{sec:objectsegmentation}

Our structurally decomposed feature field enables physically oriented segmentation and editing of the scene. We segment the scene as follows. First, we select a rendered feature component $F_{comp}(x, y)$, where $(x, y)$ is the pixel location, and $F_{comp}$ can be $F_{indep}$ or $F_{ref}$. Then, we correlate the selected feature with the corresponding decomposed feature field in 3D. The location of features with a correlation factor above a threshold belongs to the segmented region of interest in the scene. Once we obtain the region of interest, we can control and modify the physical properties of the scene \textit{locally}, such as the view-independent color, roughness, level of reflection, and more, as demonstrated in Sec.~\ref{sec:results} %

\subsection{How is Disentanglement Achieved?}

Two MLPs, $\mathbf{f}_{indep}$ and $\mathbf{f}_{dep}$, predict the view-independent and view-dependent 3D feature components (see Fig.~\ref{fig:pipeline}). To achieve the disentanglement, we enforce three constraints: (i). $\mathbf{f}_{indep}$ cannot model the view-dependent feature component. This is achieved by design, by \textbf{not} providing $\mathbf{f}_{indep}$ the view direction as input (unlike $\mathbf{f}_{dep}$). (ii). $\mathbf{f}_{indep}$ and $\mathbf{f}_{dep}$ together model the total feature value. This is enforced by summing the outputs of $\mathbf{f}_{indep}$ and $\mathbf{f}_{dep}$ together (Eq.~\ref{eq:learnable_weights_2}) and using volumetric rendering to ensure 2D rendered features match ground truth rendered features. (iii). $\mathbf{f}_{indep}$ models the view-independent feature component. While this is not explicitly enforced, we hypothesize that it is implicitly enforced by the model as it tries to be efficient (compressed). As $\mathbf{f}_{indep}$ does not utilize the view direction as input, for a given 3D position, it uses a single feature value for all view directions. $\mathbf{f}_{dep}$ can then model the residual view-dependent component only when this is \textbf{required}. For directions without reflections, for instance, this is not required, resulting in $\mathbf{f}_{dep}$ storing fewer feature values in total.

\begin{figure}
\centering
\begin{tabular}{c@{}c@{}c@{}c@{}c}
\hspace{-0.3cm}
\rotatebox{90}{\hspace{0.2cm} \tiny{Input}} %
\includegraphics[width=0.19\linewidth]{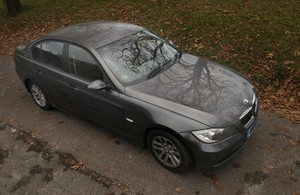} &
\includegraphics[width=0.19\linewidth]{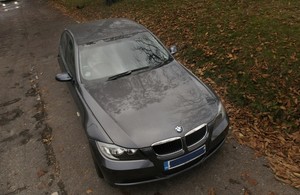} &
\includegraphics[width=0.19\linewidth]{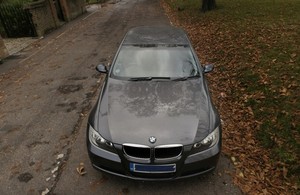} &
\includegraphics[width=0.19\linewidth]{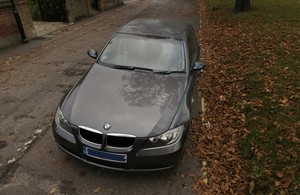} & 
\includegraphics[width=0.19\linewidth]{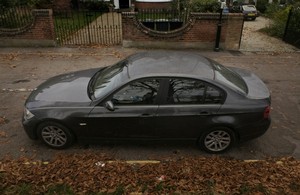} \\
\hspace{-0.3cm}
\rotatebox{90}{\hspace{0.2cm} \tiny{PCA}} %
\includegraphics[width=0.19\linewidth]{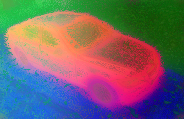} &
\includegraphics[width=0.19\linewidth]{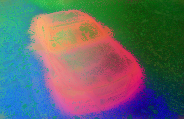} &
\includegraphics[width=0.19\linewidth]{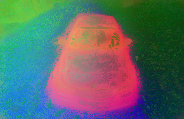} &
\includegraphics[width=0.19\linewidth]{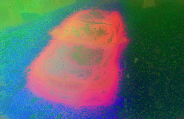} & 
\includegraphics[width=0.19\linewidth]{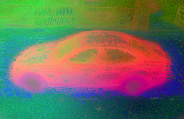} \\
\hspace{-0.3cm}
\rotatebox{90}{\hspace{0.2cm} \tiny{Zoom}} %
\includegraphics[trim={2cm 2.2cm 2cm 0.4cm},clip, width=0.19\linewidth]{figures/gt_pca/sedan/_DSC1567_feat_pca.png} &
\includegraphics[trim={1.5cm 2.2cm 2.5cm 0.4cm}, clip,width=0.19\linewidth]{figures/gt_pca/sedan/_DSC1569_feat_pca.png} &
\includegraphics[trim={2.1cm 2cm 1.9cm 0.6cm}, clip, width=0.19\linewidth]{figures/gt_pca/sedan/_DSC1570_feat_pca.png} &
\includegraphics[trim={1.9cm 2.2cm 2.1cm 0.4cm}, clip, width=0.19\linewidth]{figures/gt_pca/sedan/_DSC1571_feat_pca.png} & 
\includegraphics[trim={1.5cm 1.7cm 2.5cm 0.9cm}, clip, width=0.19\linewidth]{figures/gt_pca/sedan/_DSC1579_feat_pca.png} \\
\end{tabular}
\vspace{-0.4cm}
\caption{PCA of DINOv2 features for ground-truth input views of the Sedan scene from real-world dataset of \cite{verbin2022refnerf}. We zoom in on the windshield, illustrating differences in corresponding locations between views.}
\label{fig:gt_pca}
\vspace{-0.3cm}
\end{figure}

\section{Experiments} \label{sec:results}

We evaluate our representation on multiple vectors. 
First, we demonstrate that input features from DINOv2 contain both view-dependent and view-independent information. 
Second, we consider the ability to segment objects in 3D space. Unlike previous methods, our disentangled feature fields capture both diffuse and reflective properties and allow for a better reconstruction of the semantic components in the scene. We also enable the segmentation of the reflective component of objects in isolation from different novel views.
Third, we consider the ability to remove view-dependent (reflective) components in the scene for specific semantic objects segmented using our approach. Fourth, we consider the ability to edit the scene, where one can manipulate or change only specific objects and their dependent (reflective) properties in isolation.  

\paragraph{Datasets.} We evaluate our method on synthetic scenes from the Shiny Blender \cite{verbin2022refnerf} dataset as well as real-world scenes from the RefNeRF real-world \cite{verbin2022refnerf} dataset and Mip-NeRF-360 dataset \cite{barron2022mipnerf360}. We consider a diverse set of scenes and objects from both real world and synthetic scenes, featuring multiple objects and variable lighting conditions, highlighting the generality of our approach. 
In particular, we consider 11 scenes and 25 objects from 3 real-world and synthetic datasets. This is comparable to recent work such as DFF or RefNeRF.  We consider the standard train-view/novel-view splits provided by the respective datasets and evaluate our model on such novel views. 
Additionally, our method supports incorporating arbitrary 2D semantic features extracted from models like CLIP-Lseg and DINOv2. 
The webpage provides associated videos depicting novel views and corresponding segmentation, removal, and editing results. We also provide implementation details in the appendix. 

\begin{figure}[ht!]
\centering
\begin{tabular}{c@{}c@{}c@{}c}
Input & Ours & DFF & DFF+Ref  \\
\includegraphics[width=0.24\linewidth]{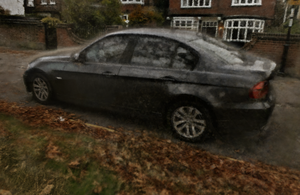} &
\includegraphics[width=0.24\linewidth]
{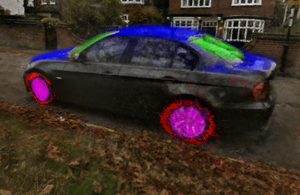} &
\includegraphics[width=0.24\linewidth]
{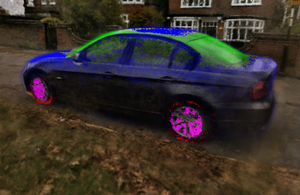} &
\includegraphics[width=0.24\linewidth]
{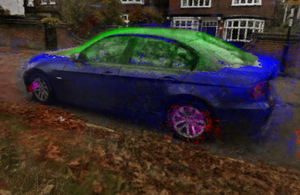} \\ 
\includegraphics[width=0.24\linewidth]{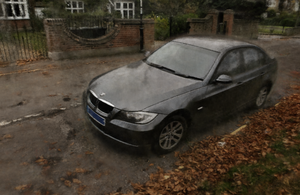} &
\includegraphics[width=0.24\linewidth]
{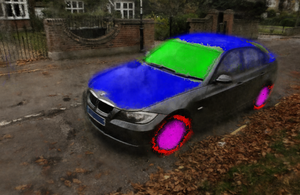} &
\includegraphics[width=0.24\linewidth]
{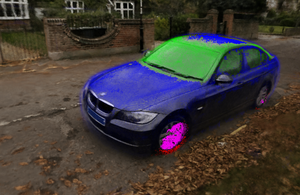} &
\includegraphics[width=0.24\linewidth]
{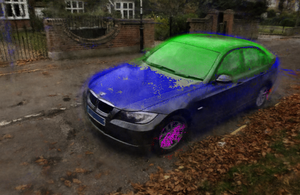} \\ 
\includegraphics[width=0.24\linewidth]{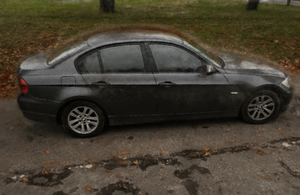} &
\includegraphics[width=0.24\linewidth]
{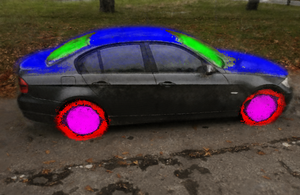} &
\includegraphics[width=0.24\linewidth]
{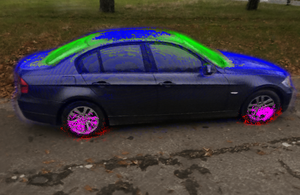} & 
\includegraphics[width=0.24\linewidth]
{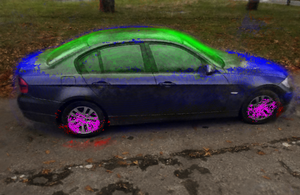} \\ 

\includegraphics[trim={0cm 0.6cm 0cm 2cm},clip, width=0.24\linewidth]
{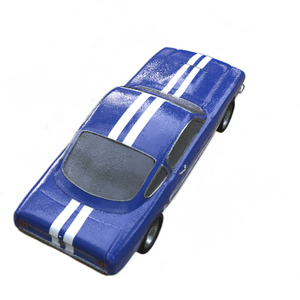} &
\includegraphics[trim={0cm 0.6cm 0cm 2cm},clip, width=0.24\linewidth]
{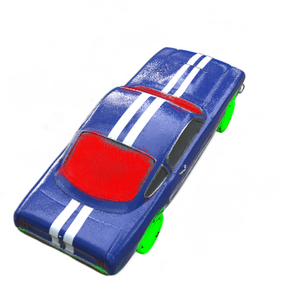} &
\includegraphics[trim={0cm 0.6cm 0cm 2cm},clip, width=0.24\linewidth]
{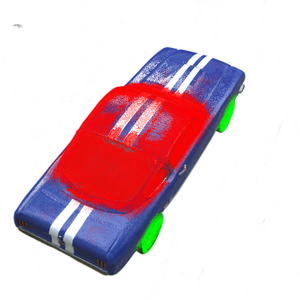} &
\includegraphics[trim={0cm 0.6cm 0cm 2cm},clip, width=0.24\linewidth]
{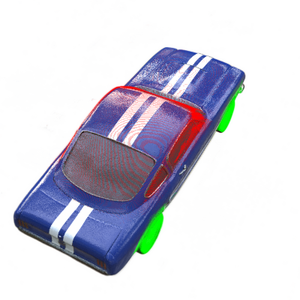} \\
\includegraphics[trim={0cm 0.6cm 0cm 2cm},clip, width=0.24\linewidth]
{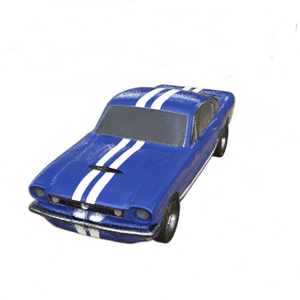} &
\includegraphics[trim={0cm 0.6cm 0cm 2cm},clip, width=0.24\linewidth]
{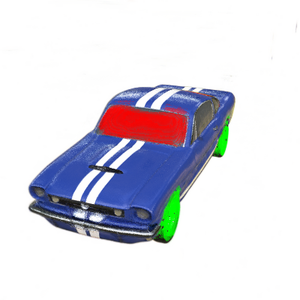} &
\includegraphics[trim={0cm 0.6cm 0cm 2cm},clip, width=0.24\linewidth]
{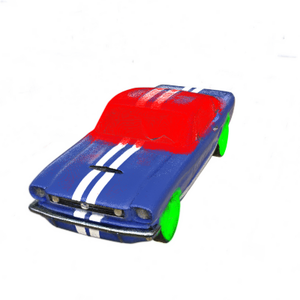} &
\includegraphics[trim={0cm 0.6cm 0cm 2cm},clip, width=0.24\linewidth]
{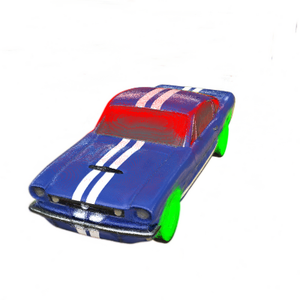} \\
\includegraphics[trim={0cm 0.6cm 0cm 2cm},clip, width=0.24\linewidth]
{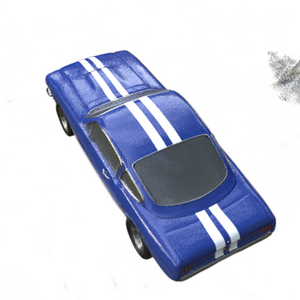} &
\includegraphics[trim={0cm 0.6cm 0cm 2cm},clip, width=0.24\linewidth]
{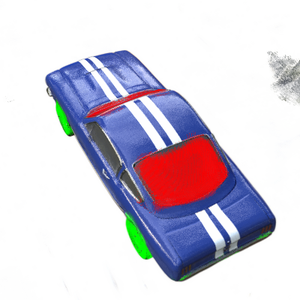} &
\includegraphics[trim={0cm 0.6cm 0cm 2cm},clip, width=0.24\linewidth]
{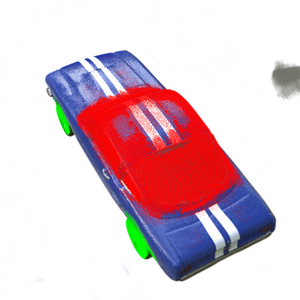} &
\includegraphics[trim={0cm 0.6cm 0cm 2cm},clip, width=0.24\linewidth]
{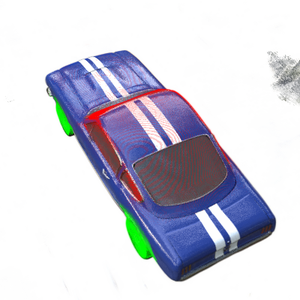} \\

\end{tabular}
\vspace{-0.3cm}
\caption{3D objects segmentation from three novel views, for the Sedan scene from real-world RefNeRF \cite{verbin2022refnerf} dataset for the objects of Bonet-top, Windshield, Hubcups and Wheels, and for the Car scene from synthetic Shiny Blender \cite{verbin2022refnerf} dataset for the objects of Windshield and Wheels. We compare our result to DFF~\cite{kobayashi2022decomposing} and to a baseline where DFF is optimized for features while RefNeRF is optimized for appearance (see \cref{sec:semantic_segmentation}).} 

\label{fig:segmentation}
\vspace{-0.3cm}
\end{figure}

\begin{table}[ht]
\centering

\scalebox{0.8}{ %
\begin{tabular}{lcccc|cc}
\toprule
\multirow{2}{*}{Scene (Objects)} & \multirow{2}{*}{Ours} & Ours & Ours & Ours & \multirow{2}{*}{DFF} & DFF+  \\
&  & implicit & total & optt & & Ref \\ 
\midrule
        Bicycle (Bench, Wheels) & \textbf{0.583} & 0.381 & 0.407 & 0.444 & 0.518 & 0.504  \\ 
        Counter (Mitten, Plant,
        & \multirow{2}{*}{\textbf{0.824}} & \multirow{2}{*}{0.705} & \multirow{2}{*}{0.641} & \multirow{2}{*}{0.664} & \multirow{2}{*}{0.713} & \multirow{2}{*}{0.629} \\ 
        Tray) \\
        Garden (Ball, Plant, 
        & \multirow{2}{*}{\textbf{0.840}} & \multirow{2}{*}{0.813} & \multirow{2}{*}{0.786} & \multirow{2}{*}{0.389} & \multirow{2}{*}{0.824} & \multirow{2}{*}{0.700} \\
        Tabletop) \\
        Kitchen (Lego) & \textbf{0.871} & 0.761 & 0.631 & 0.778 & 0.856 & 0.850 \\ 
        Gardenspheres (Cone, 
        & \multirow{2}{*}{\textbf{0.825}} &  \multirow{2}{*}{0.663} & \multirow{2}{*}{0.647} & \multirow{2}{*}{0.619} & \multirow{2}{*}{0.776} & \multirow{2}{*}{0.770} \\ 
        Head, Spheres) \\
        Sedan (Bonnet-top, Wind-
        & \multirow{2}{*}{\textbf{0.643}} &  \multirow{2}{*}{0.337} & \multirow{2}{*}{0.377} & \multirow{2}{*}{0.485} & \multirow{2}{*}{0.625} & \multirow{2}{*}{0.512} \\ 
        shield, Hubcaps, Wheel) \\
        Toycar (Body, Wheel) & \textbf{0.709} & 0.499 & 0.176 & 0.283 & 0.689 & 0.654 \\ 
        Teapot (Cover) & \textbf{0.762} & 0.125 & 0.654 & 0.768 & 0.529 & 0.399 \\ 
        \midrule
        Mean (Real World Scenes) & \textbf{0.757} & 0.628 & 0.594 & 0.523 & 0.691 & 0.582\\ 
        \midrule
        Car (Windshield, Wheels) & \textbf{0.799} & 0.437 & 0.402 & 0.088 & 0.523 & 0.445 \\
        Toaster (Body, Toasts) & 0.906 & 0.844 & 0.833 & 0.839 & \textbf{0.940} & 0.787 \\ 
        Helmet (Body, Windshield) & \textbf{0.863} & 0.795 & 0.705 & 0.894 & 0.731 & 0.472 \\ 
        \midrule
        Mean (Shiny Blender) & \textbf{0.856} & 0.568 & 0.550 & 0.648 & 0.731 & 0.527 \\ 
        \bottomrule
    \end{tabular}
}

\caption{Mean IoU for segmentation of objects from the  Shiny Blender synthetic dataset~\cite{verbin2022refnerf} (bottom) and real-world scenes (top). First four scenes are taken from the real world RefNeRF~\cite{verbin2022refnerf} dataset while the other four real world scenes are from the Mip-NeRF-360 dataset. 
On the RHS, we compare our approach DFF~\cite{kobayashi2022decomposing} and a baseline where DFF is optimized for features while RefNeRF is optimized for appearance (see \cref{sec:semantic_segmentation}). On the LHS, we also consider variants of our approach: (1). Ours-implicit: Our approach, but with implicit, instead of explicit, representation of physical properties [s.a. roughness], (2). Ours-total: Our approach, but using the total features [dependent and independent] for segmentation, (3). Our-optt: Our approach, where we optimized the color and features together. For each scene, we show, in brackets, the segmented objects. 
}
\label{tab:segmentation_iou}
\vspace{-0.3cm}
\end{table}

\begin{figure}[t]
\centering
\begin{tabular}{c@{}c@{}c@{}c@{}c}
\hspace{-0.3cm}
\rotatebox{90}{\hspace{0.1cm}\tiny{Novel View}}
\includegraphics[width=0.19\linewidth]{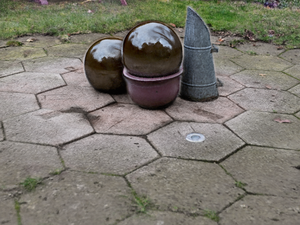} &
\includegraphics[width=0.19\linewidth]{figures/segmentation/gardenspheres/ours/no_seg/2.png} &
\includegraphics[width=0.19\linewidth]{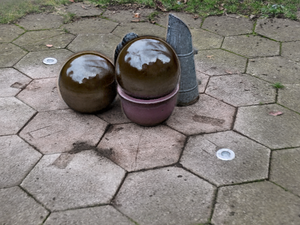} &
\includegraphics[width=0.19\linewidth]{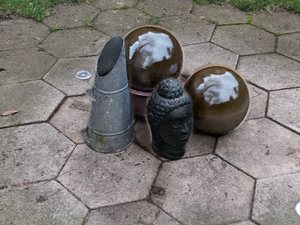} & 
\includegraphics[width=0.19\linewidth]{figures/segmentation/gardenspheres/ours/no_seg/8.png} \\

\hspace{-0.3cm}
\rotatebox{90}{\hspace{0.5cm}\tiny{Sphere} }
\includegraphics[width=0.19\linewidth]{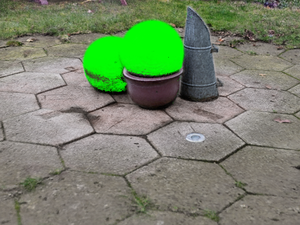} &
\includegraphics[width=0.19\linewidth]{figures/segmentation/gardenspheres/ours/spheres/2.png} &
\includegraphics[width=0.19\linewidth]{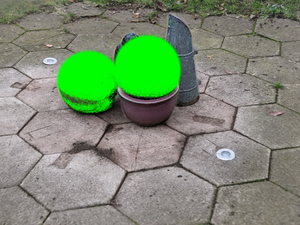} &
\includegraphics[width=0.19\linewidth]{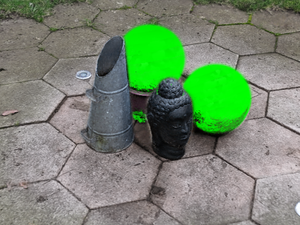} & 
\includegraphics[width=0.19\linewidth]{figures/segmentation/gardenspheres/ours/spheres/8.png} \\

\hspace{-0.3cm}
\rotatebox{90}{\hspace{0.3cm}\tiny{Reflection}}
\includegraphics[width=0.19\linewidth]{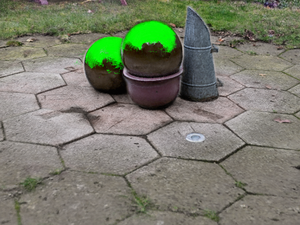} &
\includegraphics[width=0.19\linewidth]{figures/segmentation/gardenspheres/ours/highlight/2.png} &
\includegraphics[width=0.19\linewidth]{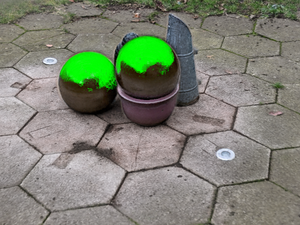} &
\includegraphics[width=0.19\linewidth]{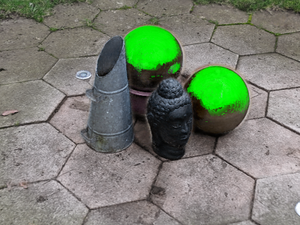} & 
\includegraphics[width=0.19\linewidth]{figures/segmentation/gardenspheres/ours/highlight/8.png} \\

\end{tabular}
\vspace{-0.3cm}
\caption{Segmentation of the spheres for novel views of the Garden-spheres real-world scene using either the full segmentation of the sphere (second row) or only the reflective component of the spheres (third row). }
\vspace{-0.3cm}
\label{fig:segmentation_combination_1}
\end{figure}

\subsection{Feature Analysis}
\label{sec:feature_analysis}

We first consider whether the distilled DINOv2 features capture both view-dependent and view-independent components. To this end, we visualize the PCA of the features for five ground-truth views of the Sedan scene from the real-world RefNeRF dataset. As seen in Fig.~\ref{fig:gt_pca}, while the features appear similar, there are notable differences, particularly in reflective regions such as the windshield (zoomed in). As further evidence, we note the recent work of \cite{el2024probing}, which demonstrates that features obtained from large foundation models, DINOv2 in particular, are not 3D view consistent. As such, applying our model has the advantage of disentangling view-dependent and view-independent components of a given feature view and can enhance downstream applications that require view-independent feature representations. This is illustrated in Sec.~\ref{sec:semantic_segmentation}, where we show that our view-independent feature field yields better 3D segmentation results than using the full (both dependent and independent) feature field or baselines.
Beyond 3D segmentation, one can render ground truth views only using the view-independent component, discarding their view-dependent component.

\subsection{Semantic Segmentation}
\label{sec:semantic_segmentation}

\begin{figure}[bh!]
\centering
\begin{tabular}{c@{}c@{}c}

\hspace{-0.3cm}
\rotatebox{90}{\hspace{0.4cm}\tiny{Novel View}}
\includegraphics[width=0.32\linewidth]{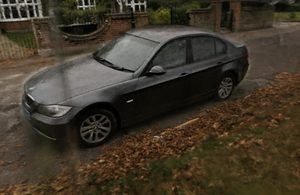} &
\includegraphics[width=0.32\linewidth]{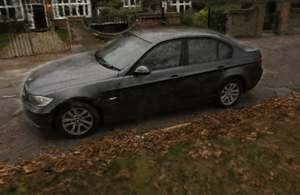} &
\includegraphics[width=0.32\linewidth]{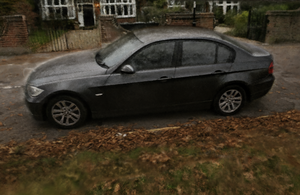} \\

\hspace{-0.3cm}
\rotatebox{90}{\hspace{0cm}\tiny{Bonet-top+W-shield}}
\includegraphics[width=0.32\linewidth]{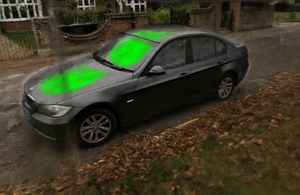} &
\includegraphics[width=0.32\linewidth]{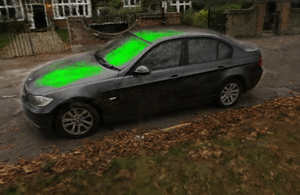} &
\includegraphics[width=0.32\linewidth]{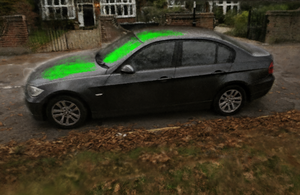} \\

\hspace{-0.3cm}
\rotatebox{90}{\hspace{0.5cm}\tiny{Bonet-top}}
\includegraphics[width=0.32\linewidth]{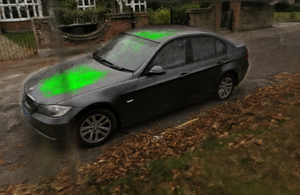} &
\includegraphics[width=0.32\linewidth]{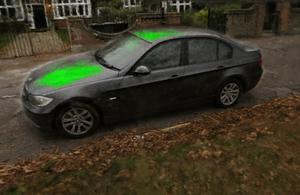} &
\includegraphics[width=0.32\linewidth]{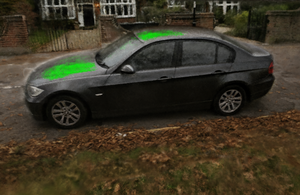} \\

\hspace{-0.3cm}
\rotatebox{90}{\hspace{0.5cm}\tiny{Windsheild}}
\includegraphics[width=0.32\linewidth]{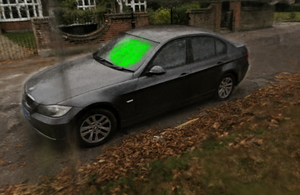} &
\includegraphics[width=0.32\linewidth]{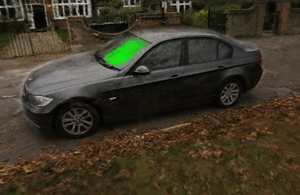} &
\includegraphics[width=0.32\linewidth]{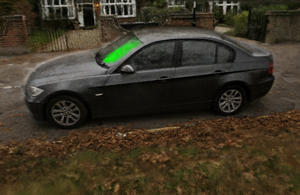} \\

\end{tabular}
\caption{
Segmentation of the reflective component of different semantic components of the real-world car scene. The first row displays three novel views. We then demonstrate the segmentation of the reflective component of (1). Both the bonnet-top and the windshield (second row), (2). The bonnet-top (third row), and (3). The windshield (fourth row).}
\label{fig:segmentation_combination_2}
\vspace{-0.4cm}

\end{figure}

We consider our method's capability in segmenting both independent and view-dependent components.

\subsubsection{Full Object Segmentation.}

In Fig.\ref{fig:segmentation}, we visualize our segmentation of three novel views for a real-world scene and a synthetic scene. The segmentation is obtained by clicking on each object and using our disentangled view-independent feature component, described in Sec.~\ref{sec:objectsegmentation}. We compare our method to Distilled Feature Field (DFF)~\cite{kobayashi2022decomposing}, the closest method to ours, which uses a single view-independent feature field. As RefNeRF is an improved appearance model, we also consider another baseline whereby appearance is obtained as in RefNeRF, using both view-dependent and view-independent feature fields, while semantics is obtained as in DFF, using a view-independent feature field. 
As can be seen, our method results in a superior segmentation and can successfully segment both reflective and non-reflective regions well, while the baseline is worse, particularly in reflective regions. Additional results are provided in the webpage.

For numerical evaluation, we compare the segmentation of objects for both synthetic scenes and real-world scenes. We manually obtain ground-truth segmentations by first applying the Segment Anything Model~\cite{kirillov2023segment} and subsequently manually refining masks (see examples in the webpage). 
As can be seen in Tab.~\ref{tab:segmentation_iou}, our method results in better mean IoU scores.

\subsubsection{Explicit modeling}
Our approach leverages explicit components to model the view-dependent and view-independent feature fields. Specifically, we model roughness and reflections explicitly, as detailed in Eq.~\ref{eq:omega_r} and Eq.~\ref{eq:reflectence_color} in Sec.~\ref{sec:method}. This enables novel applications that incorporate physical properties, such as object roughness editing. However, it is unclear whether explicit molding is preferable to implicit modeling of view-dependent and view-independent feature fields when one is interested in segmentation only. 
To evaluate the impact of explicit modeling, we considered a baseline that implicitly models view-dependent and independent feature fields by excluding the Reflection, IDE and roughness components (Eq.~\ref{eq:omega_r} and Eq.~\ref{eq:reflectence_color}). 
As seen in Tab.~\ref{tab:segmentation_iou}, this results in inferior performance.

\subsubsection{View Dependent Segmentation.}

We consider the ability to segment the view-dependent reflective surfaces of given objects. 
To this end, we begin by segmenting semantic objects using the view-independent features according to Sec.~\ref{sec:objectsegmentation} and subsequently select only a subset of points corresponding to features from the view-dependent (reflectance) feature field. 
Fig.~\ref{fig:segmentation_combination_1} illustrates for different novel views our success in segmentation for the Garden-spheres scene, for both the entire sphere as well as only the view-dependent reflection. 
Fig.~\ref{fig:segmentation_combination_2} illustrates the ability to segment the reflective region of the (1). bonnet-top and the windshield, (2). bonnet-top, and (3). windshield. 
As can be seen, only the reflective region of the desired semantic entity is depicted. 
Additional examples 
are provided in the webpage.

\subsection{Removal}

\begin{figure}
\centering
\begin{tabular}{c@{~}c@{~}c@{~}c}

\rotatebox{90}{\hspace{0.1cm} \tiny{Novel View} }
\includegraphics[width=0.21\linewidth]{figures/segmentation/gardenspheres/ours/no_seg/0.png} &
\includegraphics[width=0.21\linewidth]{figures/segmentation/gardenspheres/ours/no_seg/2.png} &
\includegraphics[width=0.21\linewidth]{figures/segmentation/gardenspheres/ours/no_seg/4.png} &
\includegraphics[width=0.21\linewidth]{figures/segmentation/gardenspheres/ours/no_seg/6.png} \\

\rotatebox{90}{\hspace{0.4cm} \tiny{Full} }
\includegraphics[width=0.21\linewidth]{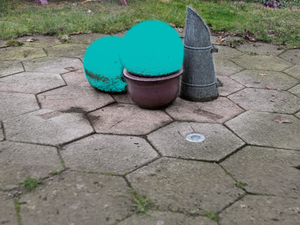} &
\includegraphics[width=0.21\linewidth]{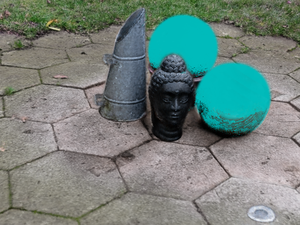} &
\includegraphics[width=0.21\linewidth]{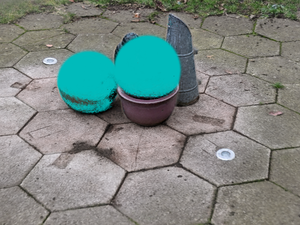} &
\includegraphics[width=0.21\linewidth]{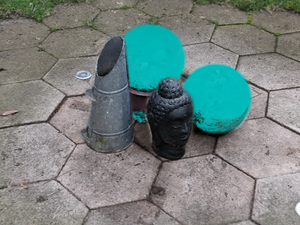} \\

\rotatebox{90}{\hspace{0.0cm} \tiny{Independent} }
\includegraphics[width=0.21\linewidth]{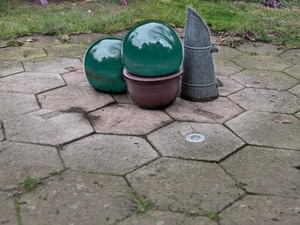} &
\includegraphics[width=0.21\linewidth]{figures/edit_new/gardenspheres/spheres/color/ours/2.png} &
\includegraphics[width=0.21\linewidth]{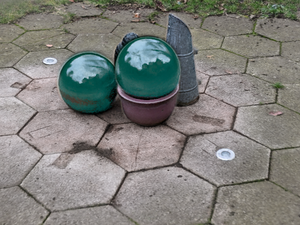} &
\includegraphics[width=0.21\linewidth]{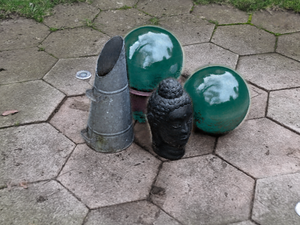}  \\

\end{tabular}

\begin{tabular}{c@{~}c@{~}c}

\rotatebox{90}{\hspace{0.7cm} \tiny{Novel View} }
\includegraphics[width=0.26\linewidth]{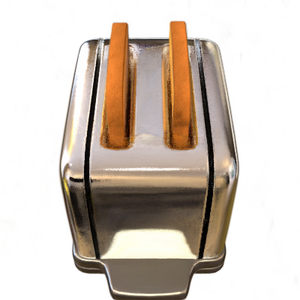} &
\includegraphics[width=0.26\linewidth]{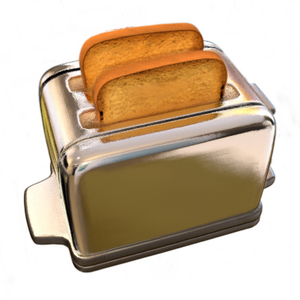} &
\includegraphics[width=0.26\linewidth]{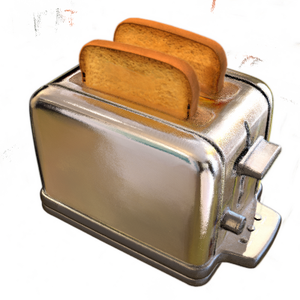} \\

\rotatebox{90}{\hspace{0.9cm} \tiny{Full} }
\includegraphics[width=0.26\linewidth]{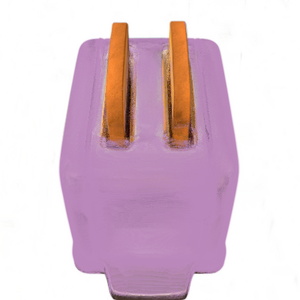} &
\includegraphics[width=0.26\linewidth]{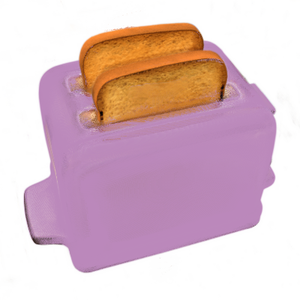} &
\includegraphics[width=0.26\linewidth]{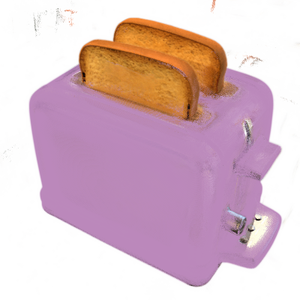} \\

\rotatebox{90}{\hspace{0.9cm} \tiny{Independent} }
\includegraphics[width=0.26\linewidth]{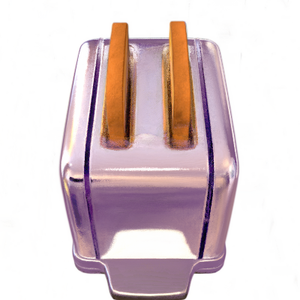} &
\includegraphics[width=0.26\linewidth]{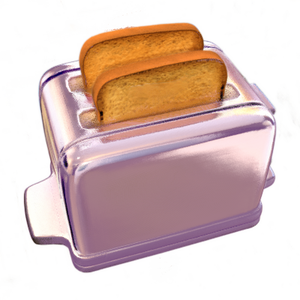} &
\includegraphics[width=0.26\linewidth]{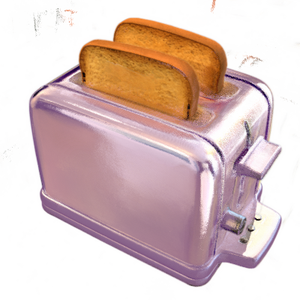}  \\

\end{tabular}
\caption{
Editing the color of i). The spheres in the real-world gardenspheresand (ii). The toaster in the synthetic toaster, either (1). using all radiance fields, (\emph{full}) or (2). using the independent component only (\emph{indepdent}). }
\label{fig:edit_color_1}
\vspace{-0.7cm}
\hspace{0.3cm}
\end{figure}

\begin{figure}[th!]
\centering
\begin{tabular}{c@{~}c@{~}c@{~}c}

\hspace{-0.3cm}
\rotatebox{90}{\hspace{0.1cm} \tiny{Novel View} }
\includegraphics[width=0.23\linewidth]{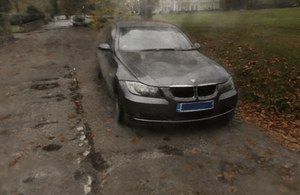} &
\includegraphics[width=0.23\linewidth]{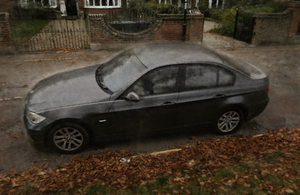} &
\includegraphics[width=0.23\linewidth]{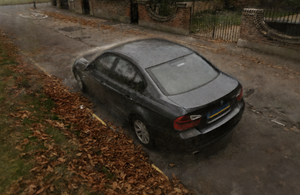} &
\includegraphics[width=0.23\linewidth]{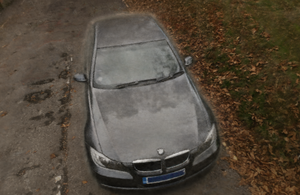} \\

\hspace{-0.3cm}
\rotatebox{90}{\hspace{0.1cm} \tiny{W/o Refl.} }
\includegraphics[width=0.23\linewidth]{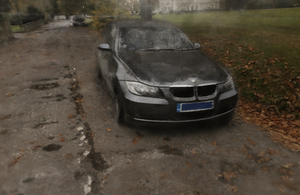} &
\includegraphics[width=0.23\linewidth]{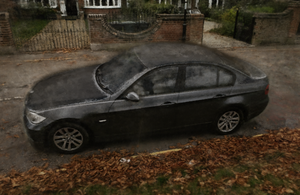} &
\includegraphics[width=0.23\linewidth]{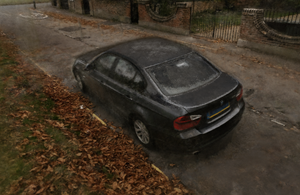} &
\includegraphics[width=0.23\linewidth]{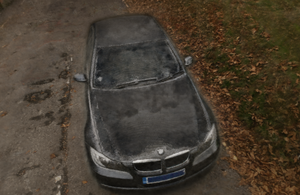} \\

\hspace{-0.3cm}
\rotatebox{90}{\hspace{0.1cm} \tiny{Novel View} }
\includegraphics[width=0.23\linewidth]{figures/segmentation/gardenspheres/ours/no_seg/0.png} &
\includegraphics[width=0.23\linewidth]{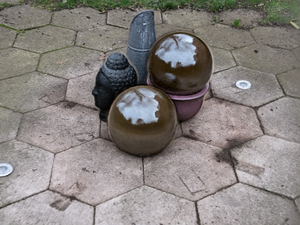} &
\includegraphics[width=0.23\linewidth]{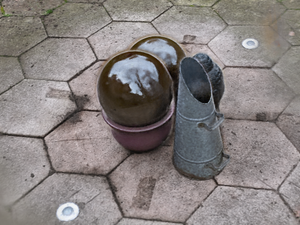} &
\includegraphics[width=0.23\linewidth]{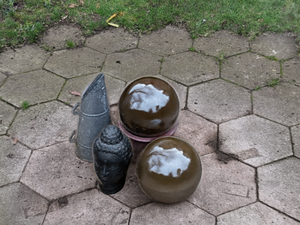} \\

\hspace{-0.3cm}
\rotatebox{90}{\hspace{0.2cm} \tiny{W/o Refl.} }
\includegraphics[width=0.23\linewidth]{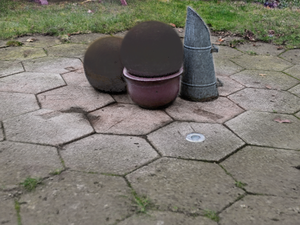} &
\includegraphics[width=0.23\linewidth]{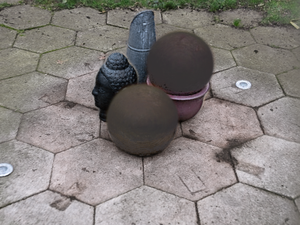} &
\includegraphics[width=0.23\linewidth]{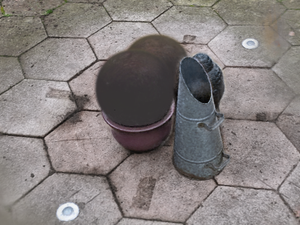} &
\includegraphics[width=0.23\linewidth]{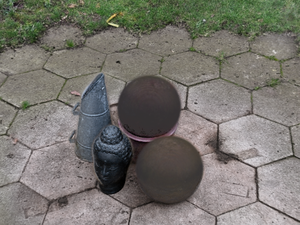} \\

\end{tabular}
\caption{
Removing the reflective component of 
of the (i). Bonet-top and windshield in the sedan scene (ii). spheres in the gardenspheres scene. 
}
\label{fig:removal_1}
\vspace{-0.2cm}
\end{figure}

Several works in the literature~\cite{mirzaei2023spin, weder2023removing, wang2023inpaintnerf360} consider the problem of 3D ``inpainting" or 3D object ``removal", where one wishes to remove a 3D foreground object, resulting in a realistic-looking background scene. Given our structurally disentangled representation, we can now explore a novel capability of ``inpainting" or ``removing" the reflective part of an object. This can be achieved by selecting the 3D points corresponding to a given object (using a click) and then rendering for those points only the color component from the view-independent radiance field.

Fig.~\ref{fig:removal_1} shows the removal of the reflective component of the bonnet-top and windshield for the car scene and the Garden-spheres scene. Our method enables the removal of the reflected radiance from the object and the retention of its diffuse color.  %

\subsection{Editing}

\subsubsection{Color Editing.}

We consider the ability to edit the color of an object while adhering to its reflections. In Fig.~\ref{fig:edit_color_1}, we change the color of the segmented 3D points of (i). the spheres, for the real-world Garden-spheres scene, and (ii). the toaster body for the synthetic toaster scene. This color change occurs either (1). using both radiance fields (independent and reflection) or (2). using the independent component only, leaving the view-dependent reflective component unchanged. Using the latter results in a more natural manipulation that correctly adheres to reflections. Additional examples are shown in the webpage. 
To assess our color editing abilities numerically, we conducted a user study on colored objects. We consider 5 colors and 5 scenes (1 object each) and asked users to assess: 1. Color faithfulness (``how well does the desired color match the object?"), 2. Realism (``how realistic does object look?"), 3. Reflections match the unedited scene (``how well do the reflections match the unedited scene?"). We considered 25 users and obtained a MOS score (1-worse, 5-best) of \textbf{4.6}/\textbf{3.1}/\textbf{4.6} vs. 2.0/2.8/\textbf{4.6} in comparison to DFF for questions 1/2/3 respectively. 

\subsubsection{Roughness Editing.}

Next, we consider the ability to manipulate physical components introduced by our architectural design. In Eq.~\ref{eq:reflectence_color}, we utilize the roughness parameter $\kappa$ that controls the roughness. To this end, we consider the ability to manipulate the roughness of individual objects in the scene. We do so by segmenting the 3D points of an object and varying the roughness parameter $\kappa$ for those points. Fig.~\ref{fig:roughness_2} illustrates two examples: (i). the spheres in the real-world scene of the Garden-spheres scene, and (ii). the helmet case for the synthetic helmet scene.  
Additional examples are provided in the webpage.

\subsubsection{Ablation Study.}

In Fig.~\ref{fig:indep_ablation}, we consider an ablation in which our segmentation is performed using not only the independent feature component $\mathbf{f}_{indep}$, but also the independent and dependent components together, $\mathbf{f}_{indep} + \mathbf{f}_{ref}$. As can be seen, this results in a worse segmentation. 

We also consider two additional ablations: (1) We optimized the appearance and feature model together, as opposed to first training the appearance model only and then the feature model (see Sec.~\ref{sec:training_objective}). For 3D segmentation (as in Tab~\ref{tab:segmentation_iou}), on average, it is worse, e.g., we get mIOU 0.648 (vs our 0.856) for Shiny Blender. (2). We also conducted an ablation where we removed the tone mapping function (Eq.~\ref{eq:learnable_weights_1}). We observe that appearance is slightly worse. Specifically, for the Gardenphere scene we obtain PSNR/LPIPS/SSIM of 28.8/0.180/0.809 vs our 28.9/0.180/0812. This results in a similar minor performance drop for 3D segmentation.

\subsubsection{Limitations.}

While our method is designed for segmenting and editing reflective regions of semantic objects in a scene, it cannot do so at the instance-based level.
For example, for the spheres in Fig.~\ref{fig:teaser}, selecting one of the spheres will result in capturing and editing both spheres. 
As in standard multiview reconstruction, errors can occur when the number of multiview ground truth features is not sufficient, or when they are noisy erroneous. 
We note that our task is highly unsupervised, as we aim to disentangle semantics and appearance in 3D space, given only 2D entangled appearance and semantics supervision. Improved physical modeling of, e.g., reflections and enhanced and generalizable feed-forward models may result in improved performance.

\section{Conclusion} \label{sec:conclusion}

We have presented a novel method for 3D scene understanding and editing by distilling pretrained 2D features into a structurally disentangled feature field representation. Our approach effectively captures both view-dependent and view-independent features, enabling superior 3D segmentation compared to prior work. We have demonstrated the unique capability of our method to segment and manipulate reflective components of objects, as well as edit object colors and roughness while preserving realistic reflections. These capabilities open up new avenues for physically based understanding and editing of 3D scenes, with potential applications in augmented reality and content creation.

In future work, beyond 3D segmentation, we aim to improve the 3D consistency of underlying foundation models such as DINOv2. To do so, one can rerender GT views only using the view-independent component, discarding their view-dependent component and then fintue DINOv2 on such features. This can enable the learning of view-independent 2D features in general, which could be useful for image/semantic correspondence. Further, we aim to extend our method to incorporate additional physical properties, such as lighting, for more comprehensive 3D scene manipulation, as well as additional features, such as those derived from text.

\clearpage
\bibliographystyle{ACM-Reference-Format}
\bibliography{references}
\clearpage
\appendix
\section{Appendix} \label{sec:appendix}

\subsection{Implementation Details}

We implement our method using Pytorch, based on the public implementation of RefNeRF~\cite{verbin2022refnerf}. In the iNGP hash grid hierarchy, we use 15 levels (from 32 to 4096) with 4 channels for each level, where each level has 4 channels. Two rounds of proposal sampling are used as in MipNeRF-360~\cite{barron2022mipnerf360}. We also penalize the mean of the sum of squared grid-hash values with a loss multiplier of 0.1. 
Our models are trained on a single A100 GPU for around 2 hours per scene. The code will also be made publically available.

\subsection{Computational requirements}
We require around 30 seconds to render a 500x400 resolution view at inference. This scales linearly with resolution. To perform segmentations and edits as detailed, we require about 60 seconds per frame on average. As for memory, we require an average of ~22GB per scene when considering high-resolution scenes with full-resolution 2D ground views and corresponding semantics. Full per-scene memory requirements will be added in the next revision.

\subsection{Per Scene IOU Results}

In \cref{tab:segmentation_iou_supp}, we provide per-scene and per-object results accompanying Tab.~1 of the main text. As can be seen, our method results in a superior performance, particularly in reflective objects such as the car windshield. 

\begin{table}[]
\centering
\begin{tabular}{lccc}
\toprule
\multirow{2}{*}{Scene - Object} & \multirow{2}{*}{Ours} &  \multirow{2}{*}{DFF} & DFF+  \\
&  & & Ref  \\ 
\midrule
Bicycle - Bench                          & \textbf{0.750}                    & 0.659                   & 0.627                       \\
Bicycle - Wheels                         & \textbf{0.415}                    & 0.378                   & 0.380                       \\
Counter - Mittens                        & 0.869                    & 0.787                   & \textbf{0.907}                       \\
Counter - Plant                          & 0.755                    & \textbf{0.791}                   & 0.256                       \\
Counter - Tray                           & \textbf{0.847}                    & 0.561                   & 0.722                       \\
Garden - Ball                            & 0.770                    & \textbf{0.838}                   & 0.732                       \\
Garden - Plant                           & \textbf{0.805}                    & 0.795                   & 0.737                       \\
Garden - Tabletop                        & \textbf{0.946}                    & 0.840                   & 0.631                       \\
Kitchen - Lego                           & \textbf{0.871}                    & 0.856                   & 0.850                       \\
Gardenspheres - Cone                     & \textbf{0.845}                    & 0.833                   & 0.817                       \\
Gardenspheres - Head                     & \textbf{0.764}                    & 0.739                   & 0.749                       \\
Gardenspheres - Spheres                  & \textbf{0.866}                    & 0.757                   & 0.742                       \\
Sedan - Bonnet-top                        & 0.471                    & 0.452                   & \textbf{0.497}                       \\
Sedan - Windshield                       & \textbf{0.659}                    & 0.569                   & 0.342                       \\
Sedan - Hubcaps                          & 0.746                    & \textbf{0.764}                   & 0.653                       \\
Sedan - Wheel                            & 0.697                    & \textbf{0.714}                   & 0.557                       \\
Toycar - Body                            & 0.775                    & 0.693                   & \textbf{0.801}                       \\
Toycar - Wheel                           & 0.644                    & \textbf{0.686}                   & 0.507                       \\
Teapot - Cover                           & \textbf{0.762}                    & 0.529                   & 0.399                       \\
\midrule
Car - Windshield & \textbf{0.838}                    & 0.357                   & 0.260                       \\
Car - Wheels     & \textbf{0.760}                    & 0.688                   & 0.630                       \\
Toaster - Body                           & 0.938                    & \textbf{0.967}                   & 0.764                       \\
Toaster - Toasts                         & 0.873                    & \textbf{0.913}                   & 0.811                       \\
Helmet - Body                            & \textbf{0.929}                    & 0.900                   & 0.860                       \\
Helmet - Windshield                      & \textbf{0.796}                    & 0.562                   & 0.084  \\                    
\bottomrule
\end{tabular}
    
\caption{Per Scene results accompanying Tab.~1 of the main text. We consider mean IoU for segmentation of objects from the Shiny Blender synthetic dataset~\cite{verbin2022refnerf} (bottom) and real-world scenes (top). The first four scenes are taken from the real-world RefNeRF~\cite{verbin2022refnerf} dataset while the other four real-world scenes are from the Mip-NeRF-360 dataset. We compare our approach to DFF~\cite{kobayashi2022decomposing} and a baseline where DFF is optimized for features while RefNeRF is optimized for appearance.
For each scene, we indicate the segmented objects. 
}
\label{tab:segmentation_iou_supp}
\vspace{-0.3cm}
\end{table}

\begin{figure}[!b]

\centering
\begin{tabular}{c@{~}c@{~}c@{~}c}

\rotatebox{90}{\hspace{0.1cm} \tiny{Novel View} }
\includegraphics[width=0.20\linewidth]{figures/segmentation/gardenspheres/ours/no_seg/0.png} &
\includegraphics[width=0.20\linewidth]{figures/segmentation/gardenspheres/ours/no_seg/2.png} &
\includegraphics[width=0.20\linewidth]{figures/segmentation/gardenspheres/ours/no_seg/4.png} &
\includegraphics[width=0.20\linewidth]{figures/segmentation/gardenspheres/ours/no_seg/6.png}\\

\rotatebox{90}{\hspace{0.3cm} \tiny{Rougher} }
\includegraphics[width=0.20\linewidth]{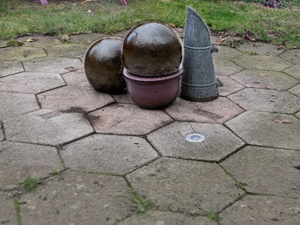} &
\includegraphics[width=0.20\linewidth]{figures/edit_new/gardenspheres/spheres/roughness/ours/2.png} &
\includegraphics[width=0.20\linewidth]{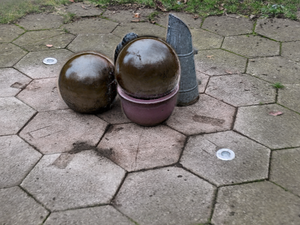} &
\includegraphics[width=0.20\linewidth]{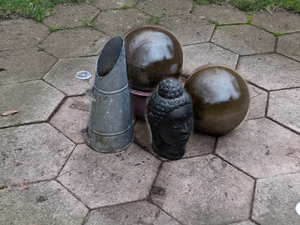} \\

\end{tabular}

\centering
\begin{tabular}{c@{~}c@{~}c@{~}c}

\rotatebox{90}{\hspace{0.5cm} \tiny{Novel View} }
\includegraphics[width=0.25\linewidth]{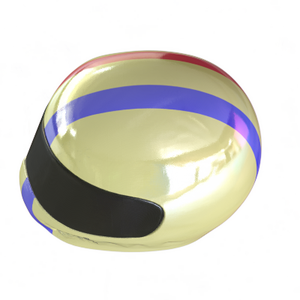} &
\includegraphics[width=0.25\linewidth]{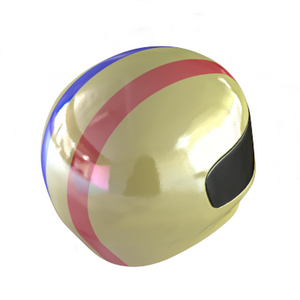} &
\includegraphics[width=0.25\linewidth]{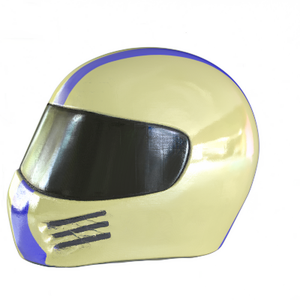} \\

\rotatebox{90}{\hspace{0.7cm} \tiny{Rougher} }
\includegraphics[width=0.25\linewidth]{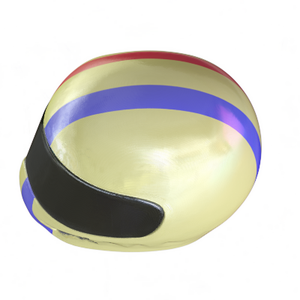} &
\includegraphics[width=0.25\linewidth]{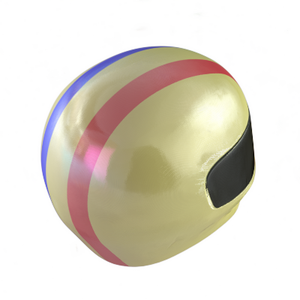} &
\includegraphics[width=0.25\linewidth]{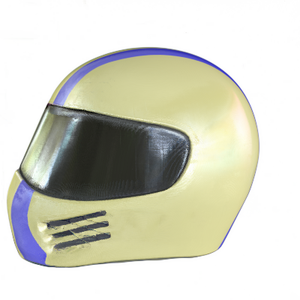} &

\end{tabular}
\caption{
Controlling the roughness of (i). The spheres in the real-world Garden-spheres, and (ii). The helmet (only the case) for the synthetic helmet scene. 
}
\vspace{-0.0cm}
\label{fig:roughness_2}
\end{figure}

\begin{figure}
\centering
\begin{tabular}{c@{~}c@{~}c@{~}c}

\hspace{-0.3cm}
\rotatebox{90}{\hspace{0.1cm} \tiny{Novel View} }
\includegraphics[width=0.23\linewidth]{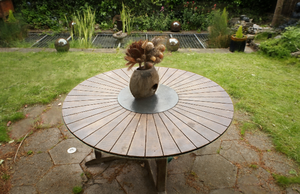} &
\includegraphics[width=0.23\linewidth]{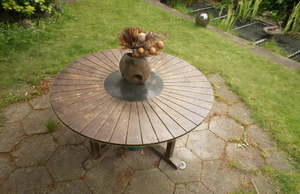} &
\includegraphics[width=0.23\linewidth]{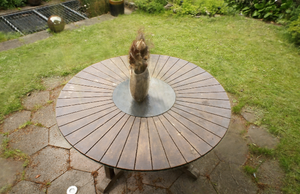} &
\includegraphics[width=0.23\linewidth]{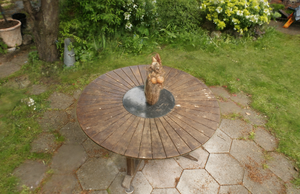} \\

\hspace{-0.3cm}
\rotatebox{90}{\hspace{0.0cm} \tiny{Indep (Ours) }}
\includegraphics[width=0.23\linewidth]{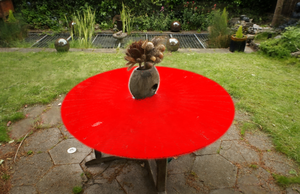} &
\includegraphics[width=0.23\linewidth]{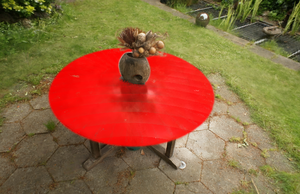} &
\includegraphics[width=0.23\linewidth]{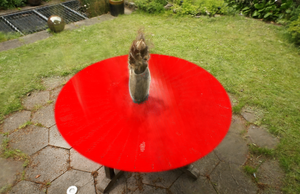} &
\includegraphics[width=0.23\linewidth]{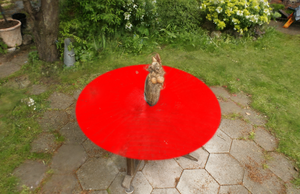} \\

\hspace{-0.3cm}
\rotatebox{90}{\hspace{0.1cm} \tiny{Indep+dep }}
\includegraphics[width=0.23\linewidth]{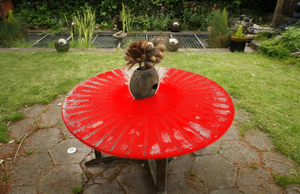} &
\includegraphics[width=0.23\linewidth]{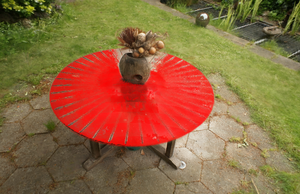} &
\includegraphics[width=0.23\linewidth]{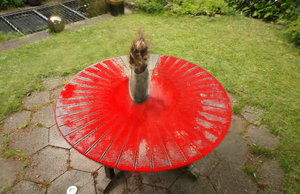} &
\includegraphics[width=0.23\linewidth]{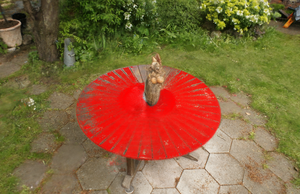} \\

\end{tabular}
\vspace{-0.3cm}
\caption{
An ablation study on 3D semantic segmentation from four novel views for the real world garden scene, comparing the segmentation obtained by clicking on each object and using (i). our view-independent feature component (used as default), or (ii) using the combination of both the view-independent and view-dependent components. 
}
\label{fig:indep_ablation}
\vspace{-0.2cm}
\end{figure}

\clearpage

\clearpage

\end{document}